\documentclass{article}

\usepackage[final]{neurips_2025}

\usepackage[utf8]{inputenc} % allow utf-8 input
\usepackage[T1]{fontenc}    % use 8-bit T1 fonts
\usepackage{hyperref}       % hyperlinks
\usepackage{url}            % simple URL typesetting
\usepackage{booktabs}       % professional-quality tables
\usepackage{amsfonts}       % blackboard math symbols
\usepackage{nicefrac}       % compact symbols for 1/2, etc.
\usepackage{microtype}      % microtypography
\usepackage[table]{xcolor}         % colors
\definecolor{lightblue}{RGB}{135, 206, 235}
\usepackage{graphicx}

\usepackage{dsfont}
\usepackage{amsmath}
\usepackage{amssymb}
\usepackage{amsthm}
\usepackage{adjustbox}
\usepackage{algorithm}
\usepackage{algorithmic}
\usepackage{wrapfig}
\usepackage{lipsum}
\usepackage{multicol}
\usepackage{subcaption}

\newtheorem{proposition}{Proposition}

\title{Retrosynthesis Planning via Worst-path Policy Optimisation in Tree-structured MDPs}

\author{%
  Mianchu Wang$^*$ \hspace{10pt} Giovanni Montana$^{*\dagger}$ \\
  $^*$University of Warwick \\
  $^\dagger$The Alan Turing Institute \\
  \texttt{\{mianchu.wang, g.montana\}@warwick.ac.uk} \\
}

\begin{document}

\maketitle

\begin{abstract}

Retrosynthesis planning aims to decompose target molecules into available building blocks, forming a synthetic tree where each internal node represents an intermediate compound and each leaf ideally corresponds to a purchasable reactant. However, this tree becomes invalid if any leaf node is not a valid building block, making the planning process vulnerable to the ``weakest link'' in the synthetic route. Existing methods often optimise for average performance across branches, failing to account for this worst-case sensitivity. 
In this paper, we reframe retrosynthesis as a worst-path optimisation problem within tree-structured Markov Decision Processes (MDPs). We prove that this formulation admits a unique optimal solution and provides monotonic improvement guarantees. Building on this insight, we introduce Interactive Retrosynthesis Planning (InterRetro), a method that interacts with the tree MDP, learns a value function for worst-path outcomes, and improves its policy through self-imitation, preferentially reinforcing past decisions with high estimated advantage. 
Empirically, InterRetro achieves state-of-the-art results -- solving 100\% of targets on the Retro*-190 benchmark, shortening synthetic routes by 4.9\%, and achieving promising performance using only 10\% of the training data.

\end{abstract}

% TLDR: We propose Interactive Retrosynthesis Planning, a novel framework that learns to construct retrosynthetic routes by interacting with tree MDPs and optimising a worst-path objective by self-imitation learning.

\section{Introduction}

\begin{wrapfigure}{r}{0.4\textwidth}
    \centering
    \footnotesize
    \includegraphics[width=0.95\linewidth]{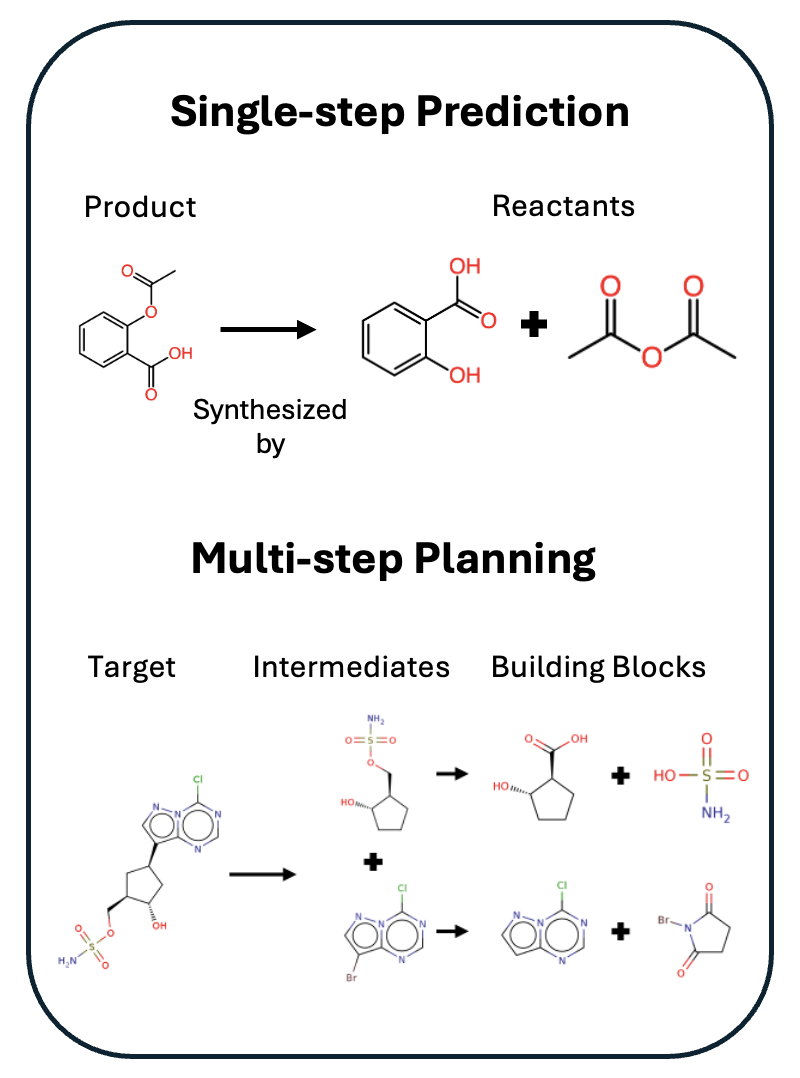}
    \caption{Single-step prediction decomposes a molecule into reactants, whereas multi-step planning searches for a synthetic route, aiming to reach purchasable building blocks.}
    \label{fig:retrosynthesis_problems}
\end{wrapfigure}

Retrosynthesis aims to identify the reactants needed to synthesise a target molecule with desired properties. As a fundamental task in computer-aided molecular design, retrosynthesis underpins progress in drug discovery and materials science \cite{dong2021deep, zhong2024recentadvances}. A key challenge in retrosynthesis is single-step prediction, which involves predicting the reactants for a given product (illustrated in Figure \ref{fig:retrosynthesis_problems}). While recent approaches using supervised learning have achieved human-level accuracy in this task \cite{chen2021localretro, zhong2023graph2edits}, their practical applicability is limited, as the suggested reactants are often commercially unavailable. Unlike single-step prediction, multi-step planning recursively decomposes the target into simpler intermediates, aiming to construct a synthetic route whose leaf nodes are purchasable compounds. This process naturally forms a sequential decision-making problem, where early decisions influence future steps and the overall outcome \cite{hong2023egmcts, schreck2019learning}.

Most methods tackle this decision-making problem using heuristic search, with Monte Carlo Tree Search (MCTS) being widely adopted for its ability to balance exploration and exploitation \cite{roucairol2024comparing, segler2018planning}. In this framework, a single-step retrosynthesis model suggests candidate reactions and predicts the most likely reactants for a given product. During simulations, MCTS selects reactions by weighing their estimated value against their exploration potential. After simulating a complete synthetic route, the values of the decisions made along the path are updated to guide future searches more effectively. Building on this foundation, recent studies have proposed various enhancements to improve the exploration--exploitation trade-off or to stabilise value estimation \cite{chen2020retrostar, hong2023egmcts, xie2022retrograph, zhao2024meea}. However, these methods heavily rely on time-consuming real-time search, requiring hundreds of model calls for each molecule, limiting their practical utility in large-scale molecular design scenarios.

To improve search efficiency, recent work fine-tunes pre-trained single-step models by imitating decisions made during successful heuristic search trajectories \cite{kim2021retrostarplus, liu2023pdvn, yu2022grasp}. The key insight is that reaction choices selected after look-ahead planning are often more reliable than those proposed by the original model \cite{schrittwieser2021reanalysis, wang2024goplan}. By training on these improved decisions, the resulting policies can propose more plausible reactions and accelerate the search for viable synthetic routes. However, this fine-tuning process adapts the model to the distribution of molecules encountered during search, rather than those seen in direct inference. As a result, the model may perform well when used within the search loop but struggle to construct full synthetic routes independently, limiting its utility in settings where fast, search-free planning is desired.

In this paper, we propose a novel perspective: reframing retrosynthesis planning as a worst-path optimisation problem in tree-structured Markov Decision Processes (tree MDPs). We observe that existing approaches typically optimise for average or cumulative rewards across all root-to-leaf paths in the synthetic tree \cite{liu2023pdvn, schreck2019learning}, overlooking a critical insight: a synthetic route is only valid if every root-to-leaf path terminates at a purchasable compound. Even a single unsuccessful path invalidates the entire synthetic route, making the worst-performing path the limiting factor in overall performance. This observation leads us to introduce a new ``worst-path'' objective that focuses explicitly on improving the most challenging path in the synthetic route.

Building on this novel formulation, we develop Interactive Retrosynthesis Planning (InterRetro) --- a framework for multi-step retrosynthesis that learns to generate complete synthetic routes without search at inference time. InterRetro treats a single-step model as an agent operating within the tree MDP, recursively decomposing molecules into reactants through environment interactions. The agent is improved via weighted self-imitation of its past successful decisions to encourage shallow synthetic trees, effectively bootstrapping its performance. In particular, InterRetro operates in three key steps. Firstly, the agent interacts with the tree MDP to construct complete synthetic routes. Secondly, it identifies successful subtrees whose leaf nodes correspond to commercially available compounds. Finally, it fine-tunes the policy to imitate decisions from these subtrees, with theoretical guarantees of monotonic improvement and convergence to a unique optimal value function. This iterative self-improvement allows InterRetro to eliminate the need for real-time search while maintaining high planning quality.

This paper makes four key contributions:
(1) we introduce the novel worst-path optimisation framework for tree-structured MDPs, specifically designed for problems where the weakest component determines the overall success;
(2) we develop a weighted self-imitation learning algorithm with monotonic improvement guarantees for optimising this worst-path objective, proving the existence of a unique optimal solution through Bellman optimality analysis;
(3) we apply this framework to retrosynthesis through InterRetro, a search-free approach to multi-step planning; and
(4) we empirically demonstrate that InterRetro outperforms state-of-the-art (SOTA) methods in terms of success rate (achieving up to 100\% on benchmark datasets), route length (reducing steps by 4.9\%), and sample efficiency (reaching 92\% of full performance with only 10\% of training data).

\section{Related Work}

\textbf{Single-step Prediction.} The fundamental task in retrosynthesis is single-step prediction, which identifies the reactants that produce a given target molecule.
Early approaches are template-based, where reaction templates are extracted from patents and literature data \cite{schneider2016USPTO}, and a model selects the most appropriate template to decompose a molecule \cite{chen2021localretro, coley2017retrosim}.
Later methods propose template-free strategies, directly mapping the SMILES string of the target molecule to that of the reactants without relying on predefined transformation patterns \cite{irwin2022chemformer, tu2022graph2smiles}.
Recent studies introduce semi-template-based approaches, which combine both paradigms by first identifying intermediate structures and then completing them into full reactants --- either by generating leaving groups \cite{somnath2021graphretro}, SMILES tokens \cite{han2024editretro, yan2020retroxpert}, or graph edits \cite{wang2023retroexplainer, zhong2023graph2edits}.
While these single-step methods achieve high accuracy in predicting known reactions, they typically optimise for likelihood of historical reactions rather than synthesisability from commercially available compounds. In contrast, our approach explicitly favours reactions that lead to purchasable building blocks, enhancing downstream planning efficiency and practical applicability.

\textbf{Multi-step Planning.} Multi-step planning aims to construct a synthetic tree rooted at the target molecule, expanded by a single-step model, and terminating at commercially available compounds. Existing methods rely on heuristic search to find viable synthetic routes. For example, \cite{heifets2021construction} formulates the retrosynthesis problem as an AND/OR tree search and adopts proof-number search to find the optimal solution; Retro* \cite{chen2020retrostar} explores the AND/OR tree using A* search and proposes value estimation methods in this setting; To improve exploration, \cite{segler2018planning} and \cite{zhao2024meea} adopt MCTS, which better balances exploitation and exploration compared to A*. These search-based approaches, however, require extensive computation at inference time, often necessitating hundreds of model calls per molecule. Recent efforts such as \cite{kim2021retrostarplus} and \cite{liu2023pdvn} address this limitation by fine-tuning the single-step model using successful trajectories collected during the planning phase, reducing search iterations and improving synthesis success rates. Yet, these methods still ultimately rely on search during inference, merely reducing rather than eliminating this computational burden.

\textbf{Self-imitation Learning.} 
Self-imitation learning improves policy performance by encouraging the agent to replicate its own high-return past behaviours \cite{oh2018selfimitation, peng2019awr}. A central aspect of self-imitation is the use of support constraints — the learned policy is regularised to remain close to the data-generating policy, which stabilises training and mitigates distributional shift \cite{peng2019awr, wang2018marwil}. This principle has been effectively applied in offline and offline-to-online reinforcement learning \cite{nair2021awac, siegel2020abm, wang2025lom}, where self-imitation facilitates safe policy improvement without extensive exploration. In our work, we apply self-imitation to retrosynthesis planning by initialising the single-step model to generate reactions from the dataset and refining it through its past successful decisions, ensuring that its proposed reactions remain chemically plausible throughout training while progressively favouring those that lead to commercially available building blocks.

\section{Retrosynthesis via Worst-path Optimisation in Tree MDPs}
Prior work in multi-step retrosynthesis typically minimises the total cost of a synthetic route by optimising cumulative rewards across all root-to-leaf paths in the synthetic tree. In contrast, our worst-path objective targets the weakest path, ensuring viability by requiring all paths to terminate at purchasable compounds. To efficiently optimise this criterion, we propose a weighted self-imitation algorithm that leverages successful trajectories while prioritising improvement on failure-prone paths.

\subsection{Tree-structured MDPs}
We formalise the retrosynthesis problem as a tree-structured Markov Decision Process (tree MDP), denoted by $\langle \mathcal{S}, \mathcal{A}, \mathcal{T}, r, \mathcal{S}_{bb} \rangle$.
Each state $s \in \mathcal{S}$ represents a molecule, and each action $a \in \mathcal{A}$ corresponds to a chemical reaction.
Unlike standard MDPs, where each transition leads to a single successor state, a chemical reaction may decompose a molecule into multiple reactants. 
To capture this branching structure, we define the power set of all possible molecules as $2^{\mathcal{S}}$ and introduce a branching transition function
$\mathcal{T} : \mathcal{S} \times \mathcal{A} \rightarrow 2^{\mathcal{S}}$, which maps a product molecule $s$ and a reaction $a$ to a set of reactants. 
These reactants become the children of $s$ in the synthetic tree.
The reward function $r : \mathcal{S} \rightarrow \mathbb{R}$ assigns a numerical value to each molecule, and $\mathcal{S}_{bb} \subset \mathcal{S}$ denotes the set of commercially available building blocks. A synthetic tree is considered successful only if all of its leaf nodes belong to $\mathcal{S}_{bb}$. A policy $\pi : \mathcal{S} \times \mathcal{A} \rightarrow [0, 1]$ defines a probability distribution over feasible reactions for a molecule $s$, thereby determining how the synthetic tree expands. Throughout this paper, we denote a complete synthetic tree by $\tau$, and let $P(\tau)$ represent the set of all root-to-leaf paths within $\tau$. The tree MDP formulation is illustrated in Figure \ref{fig:tree_mdp}.

\subsection{Worst-path Objective} 

\begin{wrapfigure}{r}{0.4\textwidth}
    \centering
    \includegraphics[width=0.95\linewidth]{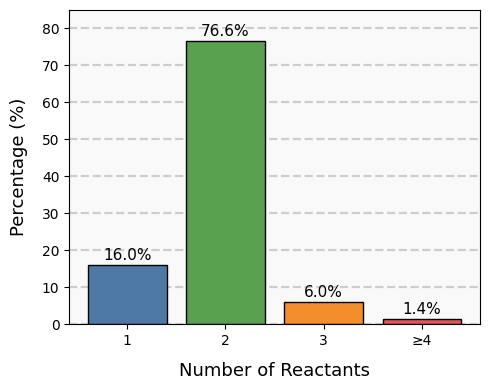}
    \caption{Distribution of reactant counts in the USPTO-50k dataset.}
    \label{fig:dist_counts}
\end{wrapfigure}

\begin{figure}[t]
    \centering
    \includegraphics[width=0.98\linewidth]{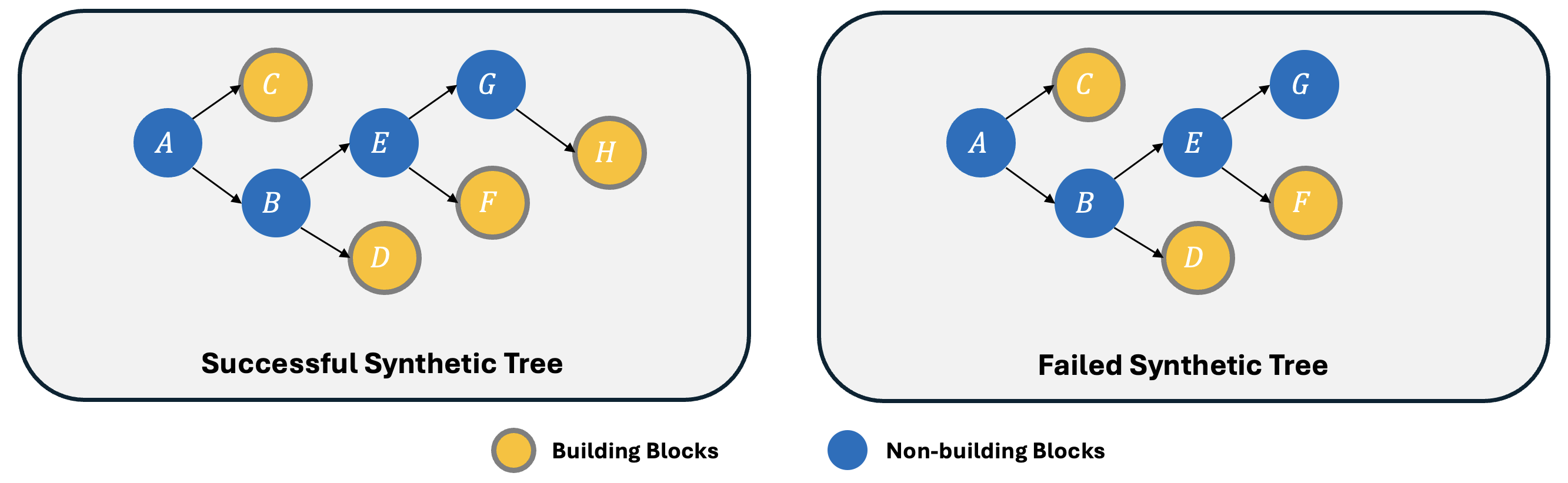} 
     \caption{Examples illustrating the tree MDP formulation. Each non-leaf node represents a molecule that is decomposed into one or more reactants. \emph{Left tree}: A successful synthetic route for target molecule $\mathrm{A}$. It contains $4$ root-to-leaf paths: $P(\tau)=\{\mathrm{ABD}, \mathrm{ABEF}, \mathrm{ABEGH}, \mathrm{AC}\}$. Since all leaf nodes are building blocks, each path receives a value of $\gamma^T$, where $T$ is the path length. The tree's overall value is $\min_{p \in P(\tau)} \{\gamma^2, \gamma^3, \gamma^4, \gamma\}=\gamma^4$, determined by the longest path. \emph{Right tree}: A failed synthesis attempt for molecule $\mathrm{A}$. One of its paths, $\mathrm{ABEG}$, terminates at $\mathrm{G}$, which is not a building block. This gives path $\mathrm{ABEG}$ a value of $0$, making the tree's overall value $\min_{p \in P(\tau)} \{\gamma^2, \gamma^3, 0, \gamma\}=0$, illustrating why a single failing path invalidates the entire route.} \label{fig:tree_mdp}
\end{figure}

The width of synthetic trees is naturally bounded, as most molecules can be synthesised from only a few reactants. 
As shown in Figure~\ref{fig:dist_counts}, $98.6\%$ of reactions in the USPTO-50k dataset involve three or fewer reactants. 
This suggests that the quality of a synthetic route is primarily determined by its depth rather than its branching factor. 
Accordingly, we evaluate each synthetic tree by the length of its longest root-to-leaf path. 
Figure~\ref{fig:tree_mdp} illustrates this concept: the left tree successfully synthesises molecule $\mathrm{A}$, with its value determined by the longest path, whereas the right tree fails due to an unsynthesisable intermediate $G$, yielding a return of $0$.

Formally, we define the reward function as
\begin{equation}
    r(s) = 
    \begin{cases}
        1, & \text{if } s \in \mathcal{S}_{bb}, \\
        0, & \text{otherwise}.
    \end{cases}
\end{equation}
This assigns a reward of $1$ to commercially available building blocks and $0$ otherwise. Since the reward is non-zero only at terminal states (i.e., leaf nodes), the return of any root-to-leaf path $p = (s_0, a_0, s_1, \ldots, s_T)$ is
\begin{equation}
  \sum_{t=0}^{T} \gamma^t r(s_t)
  = \gamma^T r(s_T),
\end{equation}
where $\gamma \in (0,1)$ is the discount factor. Hence, a successful path with $r(s_T)=1$ receives value $\gamma^T$, penalising deeper decompositions, while dead-end paths with $r(s_T)=0$ yield zero return. We then define the worst-path objective as
\begin{equation} \label{eq:worst_path_objective}
  J(\pi) = \mathbb{E}_{\tau \sim \pi}\!\left[\min_{p \in P(\tau)} \sum_{t=0}^{T} \gamma^t r(s_t)\right].
\end{equation}
This objective captures the intuition that a synthetic route is only as viable as its weakest link: if any path leads to a dead-end, the entire tree's value becomes zero. Maximising the worst-path return therefore encourages the policy to ensure that every branch terminates at building blocks through the shortest possible routes.

\subsection{Advantage Estimation}

To optimise the worst-path objective, the agent interacts with the tree-structured MDP and learns to reproduce its past advantageous decisions. In this section, we formally define the advantage to quantify how beneficial a reaction is relative to the policy’s average behaviour. We then propose to estimate this advantage using a value function learned from interaction experiences. Finally, we analyse its theoretical properties and establish the existence of an optimal policy that maximises the worst-path objective.

Given a molecule $s$ as the root node, a Q-function estimates the expected worst-path return when first expanding $s$ with reaction $a$ and subsequently following policy $\pi$:
\begin{equation}\label{eq:q_and_v}
    Q^\pi(s, a) = \mathbb{E}_{\tau \sim \pi}\!\left [ \min_{p \in P(\tau)} \sum_{t=0}^{T} \gamma^t r(s_t) \;\Big\vert\; s_0 = s, a_0 = a \right ].
\end{equation}
Here, $s_0$ and $a_0$ denote the root molecule and the initial reaction, respectively. 
Similarly, the value function $V^\pi(s)$ represents the expected worst-path return when all subsequent reactions follow policy $\pi$:
\begin{equation}\label{eq:value_fn}
    V^\pi(s) = \mathbb{E}_{\tau \sim \pi}\!\left [ \min_{p \in P(\tau)} \sum_{t=0}^{T} \gamma^t r(s_t) \;\Big\vert\; s_0 = s \right ].
\end{equation}
With these definitions in place, we can express the advantage function, which quantifies the relative benefit of applying reaction $a$ to molecule $s$ compared to following the policy:
\begin{equation}
    A^\pi(s, a) = Q^\pi(s, a) - V^\pi(s).
\end{equation}
A positive advantage indicates that reaction $a$ leads to better outcomes than the policy’s average behaviour. Proposition \ref{prop:q_and_v} establishes a relationship between $Q^\pi(s, a)$ and $V^\pi(s)$:
\begin{proposition} 
\label{prop:q_and_v}
The Q-function $Q^\pi(s, a)$ equals its immediate reward plus the discounted value of its next states:
    \begin{equation}
        Q^\pi(s, a) = r(s) + \gamma (1 - r(s)) \min_{s' \in \mathcal{T}(s, a)} V^\pi(s').
    \end{equation}
    The proof is provided in Appendix \ref{proof:q_and_v}.
\end{proposition}
Thus, we can estimate the advantage as:
\begin{equation} \label{eq:adv_estimation}
    A^\pi(s, a) = r(s) + \gamma (1 - r(s)) \min_{s' \in \mathcal{T}(s, a)} V^\pi(s')  -  V^\pi(s).
\end{equation}
Next, we derive a recursive form of the value function, which forms the foundation for learning a parameterised value function.
\begin{proposition} \label{prop:value_recursive}
    The value function $V^\pi(s)$ satisfies the recursion:
    \begin{equation}
        V^\pi(s) = r(s) + \gamma (1 - r(s)) \sum_{a \in \mathcal{A}} \pi(a \mid s) \min_{s' \in \mathcal{T}(s, a)} V^\pi(s').
    \end{equation}
    The proof is provided in Appendix \ref{proof:value_recursive}.
\end{proposition}
Beyond the value functions for a given policy, we can define the optimal worst-path value function, $V^*$, which represents the maximum possible worst-path return achievable from any state. 
\begin{proposition}\label{prop:bellman_optimality}
This optimal value function $V^*$ uniquely satisfies the Bellman optimality equation for the worst-path objective: 
\begin{equation} \label{eq:bellman_optimality}
V^*(s) = r(s) + \gamma (1 - r(s)) \max_{a \in \mathcal{A}} \left[ \min_{s' \in \mathcal{T}(s, a)} V^*(s') \right],
\end{equation}
The proof is provided in Appendix \ref{appendix:optimality_and_contraction}.
\end{proposition}
In the proof, we show that the corresponding Bellman optimality operator is a contraction mapping, which guarantees the existence and uniqueness of $V^*$ and the convergence of value iteration to $V^*$. Furthermore, there exists at least one deterministic stationary policy $\pi^*$ that is greedy with respect to $V^*$ and is therefore an optimal policy.

\subsection{Self-imitation Learning for Retrosynthesis}

A key challenge in retrosynthesis is ensuring that the learned policy proposes chemically valid reactions at each step. To address this, we constrain policy learning within the support of a pre-trained single-step model, which can empirically capture high-fidelity chemical transformations. Specifically, we leverage a pre-trained single-step model $\pi^0$, trained to reflect expert or empirical reaction knowledge from the dataset \cite{zhong2023graph2edits}, and define the constrained policy set:
\begin{equation}
    \Pi = \{ \pi \mid \pi(a \mid s) = 0 \text{ whenever } \pi^0(a \mid s) = 0 \}.
\end{equation}
By restricting $\pi$ to actions that $\pi^0$ assigns non-zero probability, we eliminate the risk of generating unrealistic reactions while allowing flexibility to re-weight feasible actions based on their effectiveness for multi-step planning.

Our objective is to find a policy $\pi \in \Pi$ that maximises the worst-path objective. We achieve this through an iterative procedure: in each iteration $i$, we aim to find an improved policy $\pi^{i+1}$ by imitating advantageous state-action pairs $(s,a)$ experienced under policy $\pi^i$. The advantage $A^{\pi^i}(s,a)$ quantifies this, and the learning objective for $\pi^{i+1}$ is formulated as:
\begin{equation} \label{eq:wil_objective}
J(\pi^{i+1}) = \mathbb{E}_{s \sim d_{\pi^i}(\cdot), a \sim \pi^i(\cdot \mid s)} \left [ \exp\bigl(\beta A^{\pi^i}(s,a)\bigr)\log \pi^{i+1}(a \mid s)\right ],
\end{equation}
where $\beta > 0$ is the advantage coefficient controlling the strength of advantage weighting, and $d_{\pi^i}$ is the state distribution induced by policy $\pi^i$ \cite{wang2018marwil}. In this case, reactions with higher advantages receive higher weights, guiding the policy toward better-than-average reactions. 

Since each new policy $\pi^{i+1}$ is derived by re-weighting $\pi^i$, and the initial policy $\pi^0$ restricts the support, the entire policy sequence $\{\pi^i\}_{i \geq 0}$ remains within the feasible set $\Pi$ \cite{mao2023str}. This ensures that all proposed reactions throughout training remain chemically valid.

\begin{proposition} \label{prop:policy_improvement}
    Let $\pi^{i+1}$ be the policy obtained by optimising the objective in Eq.\ref{eq:wil_objective}. Then, the updated policy is guaranteed to perform at least as well as the previous policy for all states:
    \begin{equation}
        V^{\pi^{i+1}}(s) \geq V^{\pi^i}(s), \forall s \in \mathcal{S}.
    \end{equation}
    The proof is provided in Appendix \ref{proof:policy_improvement}.
\end{proposition}

Proposition \ref{prop:policy_improvement} guarantees monotonic improvement under the exact policy update:
\begin{equation}
\pi^{i+1}(a \mid s) \propto \pi^i(a \mid s) \exp\bigl(\beta A^{\pi^i}(s,a)\bigr),
\end{equation}
which re-weights the previous policy by the exponentiated advantages, thereby increasing the probability of actions that yield higher returns than the current expectation.

For policy optimisation, we use the current policy $\pi^i$ to interact with the retrosynthesis environment and collect data to learn the next iteration's policy $\pi^{i+1}$. Through iterative weighted imitation, the policy increasingly favours high-quality reactions that lead to successful synthetic routes with shorter paths, while maintaining chemical plausibility by respecting the support of the pre-trained model.

\section{The InterRetro Algorithm}

\begin{algorithm*}[t]
\caption{Interactive retrosynthesis planning (InterRetro).}
\label{algo:full_algorithm}
{\setlength\multicolsep{0pt}
\begin{multicols}{2}
\begin{algorithmic}[1]
\renewcommand{\algorithmicrequire}{\textbf{Input:}}
    \REQUIRE pre-trained one-step policy $\pi_\theta$, value function $V_\phi$, training set $\mathcal{D}$, replay buffer $\mathcal{B}$.
    \renewcommand{\algorithmicrequire}{\textbf{def}}
    \REQUIRE \textsc{Explore}($\pi_\theta$, $m$):
    \STATE $\texttt{tree} \leftarrow \texttt{Tree}(\texttt{root} = m)$
    \STATE $\texttt{q} \leftarrow \{m\}$
    \STATE $\texttt{step} \leftarrow 0$
    \WHILE{$\texttt{q} \neq \emptyset$ \textbf{and} $\texttt{step} < \texttt{max\_steps}$}
        \STATE $s \leftarrow \texttt{q.pop()}$
        \STATE $a, \mathcal{S}_r \leftarrow \pi_\theta.\texttt{get\_reactants}(s)$
        \STATE $\texttt{tree.add\_branch}(s, a, \mathcal{S}_r)$
        \STATE
        \STATE \# \texttt{Add non-building blocks}
        \STATE $\texttt{q} \leftarrow \texttt{q} \cup \{\, s' \in \mathcal{S}_r \mid s' \notin \mathcal{S}_{bb} \,\}$
        \STATE
        \STATE $\texttt{step} \leftarrow \texttt{step} + 1$
    \ENDWHILE
    \RETURN $\texttt{tree}$
\end{algorithmic}
\columnbreak
\begin{algorithmic}[1]
    \renewcommand{\algorithmicrequire}{\textbf{def}}
    \REQUIRE \textsc{InterRetro}($\pi_\theta$, $m$):
    \FOR{$i = 1, \dots, I$}
    \WHILE{$\mathcal{D}$ is not empty}
        \STATE $m \leftarrow \mathcal{D}.\texttt{pop()}$
        \STATE $\texttt{tree} \leftarrow \textsc{Explore}(\pi_\theta, m)$
        \STATE $\texttt{brs} \leftarrow \{\}$
        \FOR{\textbf{each} subtree $\tau \in \texttt{tree}$}
            \IF{$\tau$ is successful}
                \STATE $\texttt{brs} \leftarrow \texttt{brs} \cup \textsc{AllBranches}(\tau)$
            \ENDIF
        \ENDFOR
        \STATE $\mathcal{B}.\texttt{append}(\texttt{brs})$
        \STATE $\texttt{branches} \leftarrow \mathcal{B}.\texttt{sample()}$
        \STATE $V_\phi.\texttt{update}(\texttt{branches})$ \makebox[2.0cm][r]{$\triangleright$ Eq.~\ref{eq:value_loss}}
        \STATE $\pi_{\theta}.\texttt{update}(V_\phi,\ \texttt{branches})$ \makebox[1.39cm][r]{$\triangleright$ Eq.~\ref{eq:policy_loss}}
    \ENDWHILE
    \ENDFOR
\end{algorithmic}
\end{multicols}}
\end{algorithm*}

We now present our algorithm InterRetro for retrosynthesis planning. Section \ref{sec:interactions} introduces how the single-step model interacts with the tree MDP and collects trajectories. Section \ref{sec:network_learning} describes how to learn a value function on the worst-path objective and how to fine-tune the policy to reproduce advantageous decisions. Finally, Section \ref{sec:implementation_details} provides more implementation details for reproducibility.

\subsection{Environment Interactions} \label{sec:interactions}

The \textsc{Explore} procedure in Algorithm \ref{algo:full_algorithm} shows how InterRetro constructs a synthetic tree by interacting with the tree MDP. Starting from a target molecule $m$, the single-step model proposes a reaction $a$ to decompose the molecule into a set of reactants $\mathcal{S}_r$. These reactants will be attached to the parent node $m$ in the synthetic tree. Among them, the non-building blocks are then placed in a collection (e.g., a queue) for further expansion. Each subsequent round pops a molecule from the collection to continue the decomposition process. The interaction terminates when the collection becomes empty (meaning all leaf nodes are building blocks) or when a predefined maximum number of steps is reached. 

\subsection{Value Function and Policy Learning} \label{sec:network_learning}

The \textsc{Explore} procedure constructs synthetic trees by interacting with the tree MDP. We extract all branches $(s, a, \mathcal{S}_r)$ within successful subtrees and store them in a first-in first-out replay buffer $\mathcal{B}$ for value function and policy learning (see \textsc{InterRetro} in Algorithm~\ref{algo:full_algorithm}). 

To learn the value function network $V_\phi$ with parameter $\phi$, we minimise the mean squared error between the predicted value $V_\phi(s)$ and its Bellman target derived from Proposition \ref{prop:value_recursive}:
\begin{equation} \label{eq:value_loss}
 \mathcal{L}(\phi) = \mathbb{E}_{(s, a, \mathcal{S}_r) \sim \mathcal{B}} [ ( V_\phi(s) - (r(s) + \gamma (1-r(s)) \min_{s' \in \mathcal{S}_r} V_{\phi^-}(s') ) )^2  ],
\end{equation}
where $V_{\phi^-}$ is a target network updated slowly for stability.

The policy network $\pi_\theta$ with pre-trained parameter $\theta$ is updated using the weighted imitation learning objective from Eq.~\ref{eq:wil_objective}, implemented as the following loss function:
\begin{equation} \label{eq:policy_loss}
\mathcal{L}(\theta) = - \mathbb{E}_{(s, a, \mathcal{S}_r)\sim\mathcal{B}} [ \exp_\mathrm{clip}(\beta A_\phi(s,a)) \log \pi_\theta (a \mid s) ],
\end{equation}
where $\beta>0$ is the advantage coefficient which controls the imitation strength on the past successful experience, and $\exp_\mathrm{clip}(\cdot)$ is the exponential function with a clipped output range $(0, C]$ for numerical stability. The advantage is estimated by the value function network, according to Eq. \ref{eq:adv_estimation}:
\begin{equation}
A_\phi(s,a) = r(s) + \gamma (1-r(s)) \min_{s' \in \mathcal{T}(s,a)} V_\phi(s') - V_\phi(s). 
\end{equation}
This joint optimisation of the value and policy networks enables InterRetro to fine-tune the single-step model, increasing the probability of reproducing high-advantage past decisions as indicated by the value function.

\subsection{Implementation Details} 
\label{sec:implementation_details}
We choose Graph2Edits as the single-step model due to its strong performance and flexibility in single-step retrosynthesis prediction \cite{zhong2023graph2edits}. Graph2Edits represents a molecule through a graph and predicts a sequence of graph edits to transform it to reactants. These graph edits can delete bonds, modify bond types, alter atoms, or attach leaving groups (see Appendix \ref{app:graph2edits} for more details). The single-step model and our proposed value function encode the molecule using a message passing neural network \cite{yang2019dmpnn} and forward the latent code into linear layers with ReLU as the activation function. For training, we run $6$ parallel exploration processes and collect $36$ synthetic trees per iteration. The networks are updated $5$ times per iteration using data from a compact replay buffer of maximal $20,000$ branches, chosen to reduce CPU memory usage and maintain close alignment between the data-collecting policy $\pi^i$ and the updated policy $\pi^{i+1}$. Our models are trained on a single NVIDIA RTX A5000 GPU and, without pre-training the single-step model, require approximately $48$ hours to fully converge. The code has been open-sourced\footnote{GitHub repository: \href{https://github.com/MianchuWang/InterRetro}{https://github.com/MianchuWang/InterRetro}.}.

\section{Experimental Results}
In this section, we aim to answer the following questions: (1) What are the advantages of our proposed method compared to the SOTA algorithms? (2) How does each component of the method contribute to the performance? Additionally, we examine the real-world feasibility of the proposed synthetic routes and present illustrative examples in Appendix \ref{sec:case_studies}.

\subsection{Benchmark Results} 
The proposed method is trained by creating synthetic routes for nearly $300k$ molecules in the USPTO-50k dataset with commercially available building blocks from the \textit{eMolecules} dataset\footnote{eMolecules: \href{https://downloads.emolecules.com/free/}{https://downloads.emolecules.com/free/}.}. We evaluate performance on three benchmarks of increasing difficulty: Retro*-190 \cite{chen2020retrostar}, ChEMBL-1000 \cite{liu2023pdvn, zdrazil2023chembl}, and GDB17-1000 \cite{liu2023pdvn, ruddigkeit2012gdb17}, where the suffix indicates the dataset size. 
We compare against established retrosynthesis planning methods: MCTS, Retro* \cite{chen2020retrostar}, Self-improve \cite{kim2021retrostarplus}, PDVN \cite{liu2023pdvn}, and GraphRetro \cite{xie2022retrograph}. Additionally, we include two single-step methods, MEGAN \cite{sacha2021megan} and Graph2Edits \cite{zhong2023graph2edits}, combined with Retro* as the search algorithm, and two recently proposed methods, DreamRetroer \cite{zhang2025dreamretroer} and RetroCaptioner \cite{liu2024retrocaptioner}. Baseline results are produced from their official implementations or the Syntheseus project \cite{maziarz2025syntheseus}.

\textbf{Success Rate.} Success rate measures the percentage of target molecules that can be successfully decomposed into building blocks. In Table \ref{tab:exp_succ_rate}, we firstly compare methods under different model-call budgets: $100$, $200$, and $500$. Our proposed InterRetro significantly outperforms SOTA algorithms across all three test sets. With $500$ model calls, InterRetro achieves $100\%$, $98.2\%$, and $99.5\%$ success rates on Retro*-190, ChEMBL-1000, and GDB17-1000, respectively. All synthetic routes generated by InterRetro on Retro*-190 using 500 model calls are provided in the supplementary materials.

\begin{table}[t]
    \centering
    \footnotesize
    \caption{Performance evaluation on three benchmarks. The evaluation metrics include the success rate under different test molecules with different budgets of model calls, which are direct generation (DG), $100$, $200$ and $500$ model calls. The DG columns are single-step model's results without search.}
    \label{tab:exp_succ_rate}
    \adjustbox{max width=\textwidth}{
    \begin{tabular}{cccccccccccccc}
    \toprule
    & & \multicolumn{4}{c}{\textbf{Retro$^*$-190}} & \multicolumn{4}{c}{\textbf{ChEMBL-1000}} & \multicolumn{4}{c}{\textbf{GDB17-1000}} \\
     \cmidrule(lr){3-6} \cmidrule(lr){7-10} \cmidrule(lr){11-14}    
    \textbf{Single-step} &\textbf{Search}  & \cellcolor{lightblue!40} DG & 100 & 200 & 500 & \cellcolor{lightblue!40} DG & 100 & 200 & 500 & \cellcolor{lightblue!40} DG & 100 & 200 & 500 \\
    \midrule
    Template & MCTS     & \cellcolor{lightblue!40} $20.00$  & $43.68$  & $47.37$  & $62.63$  & \cellcolor{lightblue!40} $32.00$ & $45.60$ & $68.80$ & $71.90$ & \cellcolor{lightblue!40} $3.00$ & $3.20$ & $3.70$ & $4.50$ \\
    Template & Retro*  & \cellcolor{lightblue!40} $20.00$ & $38.42$ & $58.42$ & $75.26$ & \cellcolor{lightblue!40} $32.00$ & $69.10$  & $72.00$ & $74.70$ & \cellcolor{lightblue!40} $3.00$ & $5.40$ & $6.60$ & $7.50$ \\
    LocalRetro & MCTS & \cellcolor{lightblue!40} 22.10 & $44.21$ & $57.36$ & $62.10$ & \cellcolor{lightblue!40} $47.30$ & $62.70$ & $69.10$ & $75.00$ & \cellcolor{lightblue!40} $4.60$ & $14.00$ & $16.70$ & $20.30$ \\
    LocalRetro & Retro* & \cellcolor{lightblue!40} $22.10$ & $58.94$ & $64.73$ & $73.68$ & \cellcolor{lightblue!40} $47.30$ & $74.80$ & $80.40$ & $82.40$ & \cellcolor{lightblue!40} $4.60$ & $18.90$ & $22.20$ & $28.80$ \\
    MEGAN & Retro* & \cellcolor{lightblue!40} $8.42$ & $60.52$ & $62.10$ & $73.15$ & \cellcolor{lightblue!40} $38.00$ & $71.70$ & $75.40$ & $79.00$ & \cellcolor{lightblue!40} $6.20$ & $37.60$ & $ 45.70$ & $57.20$ \\
    Graph2Edits & Retro* & \cellcolor{lightblue!40} $16.84$ & $41.05$ & $50.00$ & $56.31$ & \cellcolor{lightblue!40} $47.10$ & $68.70$ & $78.80$ & $80.70$  & \cellcolor{lightblue!40} $5.90$ & $18.20$ & $24.00$ & $32.20$ \\
    Self-improve & Retro* & \cellcolor{lightblue!40} $-$ & $67.37$ & $83.16$ & $94.74$ & \cellcolor{lightblue!40} $-$ & $-$ & $-$ & $81.10$ & \cellcolor{lightblue!40} $-$ & $-$ & $-$ & $15.00$ \\
    PDVN  & Retro* & \cellcolor{lightblue!40} $-$ & $93.68$ & $97.37$ & $98.95$ & \cellcolor{lightblue!40} $-$ & $-$ & $-$ & $83.50$ & \cellcolor{lightblue!40} $-$ & $-$ & $-$ & $26.90$ \\
    RetroCaptioner & Retro* & \cellcolor{lightblue!40} $5.26$ & $68.94$  & $72.63$  & $85.26$  & \cellcolor{lightblue!40} $3.90$ & $72.60$ & $76.50$ & $78.70$ & \cellcolor{lightblue!40} $3.20$ & $56.20$ & $68.20$ & $75.20$ \\
    \multicolumn{2}{c}{DreamRetroer} & \cellcolor{lightblue!40} $32.10$ & $78.94$  & $88.42$  & $90.52$  & \cellcolor{lightblue!40} $31.10$ & $78.10$ & $81.70$ & $83.10$ & \cellcolor{lightblue!40} $4.20$ & $27.36$ & $28.97$ & $33.20$ \\
    \midrule
    InterRetro & MCTS & \cellcolor{lightblue!40} $\pmb{95.78}$ & $89.47$ & $98.94$ & $\pmb{100.00}$ & \cellcolor{lightblue!40} $\pmb{93.10}$ & $78.40$   & $89.30$ & $97.50$ & \cellcolor{lightblue!40} $\pmb{89.00}$ & $80.80$ & $96.10$ & $\textbf{99.50}$\\
    InterRetro & Retro* & \cellcolor{lightblue!40} $\pmb{95.78}$ & $\pmb{96.31}$ & $\pmb{100.00}$ & $\pmb{100.00}$ & \cellcolor{lightblue!40} $\pmb{93.10}$ & $\pmb{91.40}$ & $\pmb{96.20}$ & $\pmb{98.20}$ & \cellcolor{lightblue!40} $\pmb{89.00}$ & $\pmb{83.80}$ & $\pmb{96.50}$ & $97.20$\\
    \bottomrule
    \end{tabular}
    }
\end{table}

Most notably, InterRetro maintains exceptional performance on the challenging GDB17-1000 benchmark, where prior approaches such as PDVN and Self-improve struggle ($26.9\%$ and $15.0\%$ respectively). This dramatic performance gap suggests that our worst-path objective effectively handles complex molecules that require precise reaction planning.

Furthermore, InterRetro can directly generate high-quality synthetic routes without any search algorithm, as shown in the Direct Generation (DG) columns. Our search-free performance ($95.78\%$, $93.10\%$, and $89.00\%$ across the three benchmarks) substantially exceeds even the search-based performance of competing methods. This demonstrates that our self-imitation learning approach successfully transfers planning capabilities to inference time, effectively eliminating the computational bottleneck of real-time search.

\textbf{Route Length.} Route length is the number of reactions needed to synthesise a target molecule. Route length could be a concern with our worst-path objective since it does not consider the width of the tree and the total number of reactions required. In Table \ref{tab:route_length}, we compare the route length with other baselines on the $138$ target molecules that all methods can resolve. Our method outperforms the SOTA by $4.85\%$.

\begin{wraptable}{r}{0.4\textwidth}
    \centering
    \caption{The average length of the routes on the Retro*-190 test set.}
    \label{tab:route_length}
    \begin{tabular}{cc}
        \toprule
        \textbf{Algorithm}     & \textbf{Average Length} \\
        \midrule
        Retro*     &  5.83  \\
        RetroGraph &  5.63  \\
        PDVN       &  4.83 \\
        MEGAN      & 4.12 \\
        Graph2Edits & 4.38  \\
        \midrule
        InterRetro       & \pmb{3.92}   \\
        \bottomrule
    \end{tabular}
\end{wraptable}

\begin{figure}[t]
    \centering
    \begin{subfigure}[b]{0.325\textwidth}
        \centering
        \includegraphics[width=\textwidth]{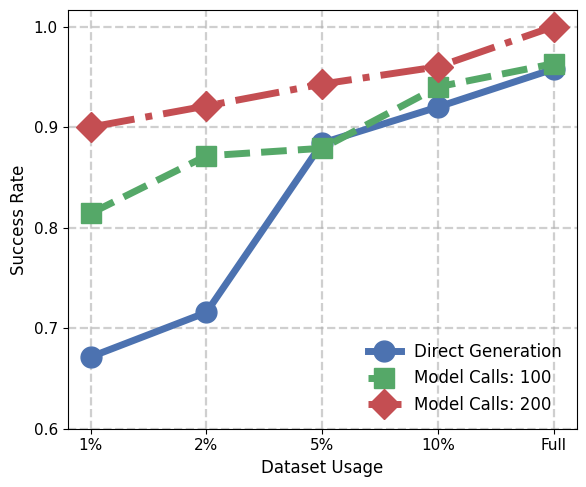}
        \subcaption{}
        \label{fig:sample_efficiency}
    \end{subfigure}
    \begin{subfigure}[b]{0.325\textwidth}
        \centering
        \includegraphics[width=\textwidth]{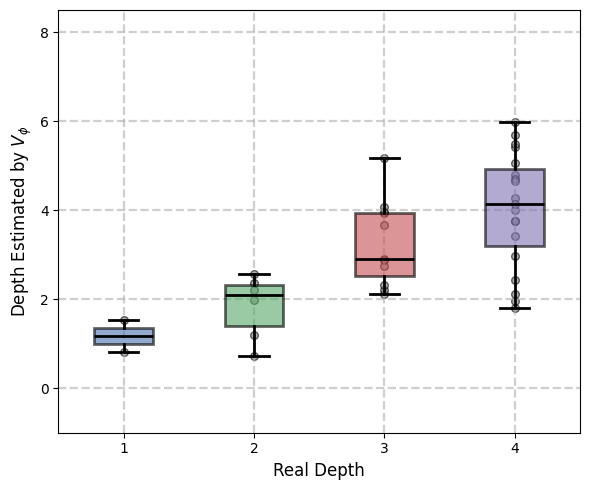}
        \subcaption{}
        \label{fig:value_estimation}
    \end{subfigure}
    \begin{subfigure}[b]{0.325\textwidth}
        \centering
        \includegraphics[width=\textwidth]{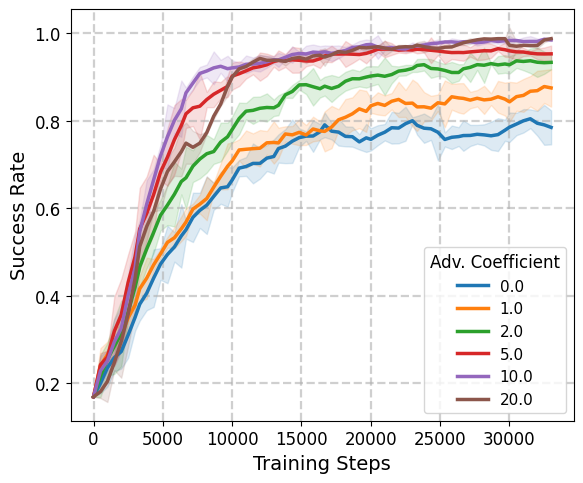}
        \subcaption{}
        \label{fig:ablation_betas}
    \end{subfigure}
    \caption{Experimental figures. \textbf{(a)} Performance under different training data usage and computation budgets. \textbf{(b)} Statistics on estimated depth of synthetic trees. \textbf{(c)} Ablations on the advantage coefficient.}
\end{figure}

\textbf{Sample Efficiency.} Sample efficiency refers to the amount of training data required to learn an effective policy. We evaluate the policy using subsets of the training data at $1\%$, $2\%$, $5\%$, $10\%$, and $100\%$, corresponding to approximately $3k$, $6k$, $15k$, $30k$, and $300k$ molecules. As shown in Figure \ref{fig:sample_efficiency}, InterRetro achieves near SOTA performance ($92\%$) with just $10\%$ of the training set when directly generating synthetic trees, and reaches SOTA performance when combined with $200$ Retro* search iterations. Notably, when trained on the full dataset, InterRetro outperforms most existing methods in direct generation, achieving a success rate of $95.78\%$. Similar trends are observed on the ChEMBL-1000 and GDB17-1000 benchmarks.

\subsection{Ablation Studies}
\textbf{Value Estimation.} We proposed a value function to estimate the worst-path return. The worst-path return indicates the estimated depth of the synthetic tree: $\text{depth}(s) = \frac{\log V_\phi(s)}{\log \gamma}$. In Figure \ref{fig:value_estimation}, we investigate the distribution of the estimated depth, corresponding to the real depth by direct generation. We found that the value function can reflect the difficulties to synthesise the molecules, while it shows better capability on the molecules that require a less deep synthetic route.

\textbf{Advantage Coefficient.} The hyperparameter $\beta$ controls the imitation strength on the past successful experience. A large $\beta$ means a focused imitation on the action with a high advantage. Figure \ref{fig:ablation_betas} shows the learning curve on the Retro*-190 test set with $\beta \in \{0, 1, 2, 5, 10, 20\}$. The figure shows that the DG performance increases from $16\%$ to around $80\%$ by uniformly imitating successful past experiences. With the advantage weighting, the performance soars to a success rate of more than $90\%$ when $\beta$ increases to $2$, and more than $95\%$ with $\beta \geq 10$.

\section{Conclusions and Discussion}
\label{sec:discussion_and_conclusions}
We introduced InterRetro, a novel approach that frames retrosynthesis planning as worst-path optimisation in tree-structured MDPs. Our weighted self-imitation algorithm enables single-step models to generate high-quality synthetic routes without search at inference time, achieving SOTA performance in success rate, route length, and sample efficiency.

Further investigation can proceed in three main directions.
First, the worst-path objective could be extended to incorporate additional real-world constraints, such as reaction conditions and the cost of building blocks. Second, and perhaps most importantly, our method and other contemporary work assume that all proposed reactions are feasible in real-world settings — an assumption that does not always hold. Although we mitigate this by imitating only actions supported by pre-trained models, computational approaches cannot guarantee practical feasibility without experimental validation. Future work can therefore focus on integrating reaction feasibility checks into InterRetro. Furthermore, the ``weakest-link'' principle underlying our method is broadly applicable beyond chemistry, extending to sequential decision problems where overall success depends on the least reliable component — for example, in robust project planning where delays in any critical task impact the entire timeline, or in multi-agent systems where team performance is constrained by the least capable agent.

\textbf{Broader Impacts.} 
The retrosynthesis community has recently open-sourced many high-quality models capable of suggesting synthetic routes for a vast number of molecules — including both beneficial drugs and potentially harmful compounds.
To promote safe and ethical research, we propose that access to such models or their source code be governed by a Responsible Use License, under which researchers acknowledge the responsible and lawful application of the technology.

\section*{Acknowledgments} 

We acknowledge support from a UKRI Turing AI Acceleration Fellowship (EPSRC EP/V024868/1). 

\bibliographystyle{plain} 
\bibliography{references}

@InProceedings{liu2023pdvn,
  title = 	 {Retrosynthetic Planning with Dual Value Networks},
  author =       {Liu, Guoqing and Xue, Di and Xie, Shufang and Xia, Yingce and Tripp, Austin and Maziarz, Krzysztof and Segler, Marwin and Qin, Tao and Zhang, Zongzhang and Liu, Tie-Yan},
  booktitle = 	 {Proceedings of the 40th International Conference on Machine Learning},
  pages = 	 {22266--22276},
  year = 	 {2023},
  volume = 	 {202},
  series = 	 {Proceedings of Machine Learning Research},
  month = 	 {Jul},
  publisher =    {PMLR},
  url = 	 {https://proceedings.mlr.press/v202/liu23as.html}
}

@ARTICLE{schreck2019learning,
  title     = "Learning Retrosynthetic Planning through Simulated Experience",
  author    = "Schreck, John S and Coley, Connor W and Bishop, Kyle J M",
  journal   = "ACS Cent. Sci.",
  publisher = "American Chemical Society",
  volume    =  5,
  number    =  6,
  pages     = "970--981",
  month     =  {Jun},
  year      =  2019
}

@InProceedings{kim2021retrostarplus,
  title = 	 {Self-Improved Retrosynthetic Planning},
  author =       {Kim, Junsu and Ahn, Sungsoo and Lee, Hankook and Shin, Jinwoo},
  booktitle = 	 {Proceedings of the 38th International Conference on Machine Learning},
  pages = 	 {5486--5495},
  year = 	 {2021},
  volume = 	 {139},
  series = 	 {Proceedings of Machine Learning Research},
  month = 	 {Jul},
  publisher =    {PMLR},
  url = 	 {https://proceedings.mlr.press/v139/kim21b.html}
}

@InProceedings{chen2020retrostar,
  title = 	 {Retro*: Learning Retrosynthetic Planning with Neural Guided {A}* Search},
  author =       {Chen, Binghong and Li, Chengtao and Dai, Hanjun and Song, Le},
  booktitle = 	 {Proceedings of the 37th International Conference on Machine Learning},
  pages = 	 {1608--1616},
  year = 	 {2020},
  volume = 	 {119},
  series = 	 {Proceedings of Machine Learning Research},
  month = 	 {Jul},
  publisher =    {PMLR},
  url = 	 {https://proceedings.mlr.press/v119/chen20k.html}
}

@article{zhong2024recentadvances,
author = {Zhong, Zipeng and Song, Jie and Feng, Zunlei and Liu, Tiantao and Jia, Lingxiang and Yao, Shaolun and Hou, Tingjun and Song, Mingli},
title = {Recent advances in deep learning for retrosynthesis},
journal = {WIREs Computational Molecular Science},
volume = {14},
number = {1},
pages = {e1694},
doi = {https://doi.org/10.1002/wcms.1694},
url = {https://wires.onlinelibrary.wiley.com/doi/abs/10.1002/wcms.1694},
eprint = {https://wires.onlinelibrary.wiley.com/doi/pdf/10.1002/wcms.1694},
year = {2024}
}

@inproceedings{xie2022retrograph,
author = {Xie, Shufang and Yan, Rui and Han, Peng and Xia, Yingce and Wu, Lijun and Guo, Chenjuan and Yang, Bin and Qin, Tao},
title = {{RetroGraph}: Retrosynthetic Planning with Graph Search},
year = {2022},
isbn = {9781450393850},
publisher = {Association for Computing Machinery},
address = {New York, NY, USA},
url = {https://doi.org/10.1145/3534678.3539446},
doi = {10.1145/3534678.3539446},
booktitle = {Proceedings of the 28th ACM SIGKDD Conference on Knowledge Discovery and Data Mining},
pages = {2120–2129},
numpages = {10},
location = {Washington DC, USA},
series = {KDD '22}
}

@ARTICLE{schneider2016uspto,
  title     = "What's What: The (Nearly) Definitive Guide to Reaction Role
               Assignment",
  author    = "Schneider, Nadine and Stiefl, Nikolaus and Landrum, Gregory A",
  journal   = "J. Chem. Inf. Model.",
  publisher = "American Chemical Society",
  volume    =  56,
  number    =  12,
  pages     = "2336--2346",
  month     =  {Dec},
  year      =  2016
}

@inproceedings{somnath2021graphretro,
 author = {Somnath, Vignesh Ram and Bunne, Charlotte and Coley, Connor and Krause, Andreas and Barzilay, Regina},
 booktitle = {Advances in Neural Information Processing Systems},
 pages = {9405--9415},
 publisher = {Curran Associates, Inc.},
 title = {Learning Graph Models for Retrosynthesis Prediction},
 url = {https://proceedings.neurips.cc/paper_files/paper/2021/file/4e2a6330465c8ffcaa696a5a16639176-Paper.pdf},
 volume = {34},
 year = {2021}
}

@article{irwin2022chemformer,
doi = {10.1088/2632-2153/ac3ffb},
url = {https://dx.doi.org/10.1088/2632-2153/ac3ffb},
year = {2022},
month = {Jan},
publisher = {IOP Publishing},
volume = {3},
number = {1},
pages = {015022},
author = {Irwin, Ross and Dimitriadis, Spyridon and He, Jiazhen and Bjerrum, Esben Jannik},
title = {Chemformer: a pre-trained transformer for computational chemistry},
journal = {Machine Learning: Science and Technology}
}

@inproceedings{yan2020retroxpert,
 author = {Yan, Chaochao and Ding, Qianggang and Zhao, Peilin and Zheng, Shuangjia and YANG, JINYU and Yu, Yang and Huang, Junzhou},
 booktitle = {Advances in Neural Information Processing Systems},
 pages = {11248--11258},
 publisher = {Curran Associates, Inc.},
 title = {{RetroXpert}: Decompose Retrosynthesis Prediction Like A Chemist},
 url = {https://proceedings.neurips.cc/paper_files/paper/2020/file/819f46e52c25763a55cc642422644317-Paper.pdf},
 volume = {33},
 year = {2020}
}

@ARTICLE{zhong2023graph2edits,
  title    = "Retrosynthesis prediction using an end-to-end graph generative
              architecture for molecular graph editing",
  author   = "Zhong, Weihe and Yang, Ziduo and Chen, Calvin Yu-Chian",
  journal  = "Nature Communications",
  volume   =  14,
  number   =  1,
  pages    = "3009",
  month    =  {May},
  year     =  2023
}

@ARTICLE{han2024editretro,
  title    = "Retrosynthesis prediction with an iterative string editing model",
  author   = "Han, Yuqiang and Xu, Xiaoyang and Hsieh, Chang-Yu and Ding, Keyan
              and Xu, Hongxia and Xu, Renjun and Hou, Tingjun and Zhang, Qiang
              and Chen, Huajun",
  journal  = "Nature Communications",
  volume   =  15,
  number   =  1,
  pages    = "6404",
  month    =  {Jul},
  year     =  2024
}

@article{heifets2021construction, 
    title={Construction of New Medicines via Game Proof Search}, 
    volume={26}, 
    url={https://ojs.aaai.org/index.php/AAAI/article/view/8331}, 
    DOI={10.1609/aaai.v26i1.8331}, 
    abstractNote={ &lt;p&gt; The production of any new medicine requires solutions to many planning problems. The most fundamental of these is determining the sequence of chemical reactions necessary to physically create the drug. Surprisingly, these organic syntheses can be modeled as branching paths in a discrete, fully-observable state space, making the construction of new medicines an application of heuristic search. We describe a model of organic chemistry that is amenable to traditional AI techniques from game tree search, regression, and automatic assembly sequencing. We demonstrate the applicability of AND/OR graph search by developing the first chemistry solver to use proof-number search. Finally, we construct a benchmark suite of organic synthesis problems collected from undergraduate organic chemistry exams, and we analyze our solvers performance both on this suite and in recreating the synthetic plan for a multibillion dollar drug. &lt;/p&gt; }, 
    number={1}, 
    journal={Proceedings of the AAAI Conference on Artificial Intelligence}, 
    author={Heifets, Abraham and Jurisica, Igor}, 
    year={2021}, 
    month={Sep.}, 
    pages={1564-1570} 
}

@ARTICLE{segler2018planning,
  title    = "Planning chemical syntheses with deep neural networks and
              symbolic {AI}",
  author   = "Segler, Marwin H S and Preuss, Mike and Waller, Mark P",
  journal  = "Nature",
  volume   =  555,
  number   =  7698,
  pages    = "604--610",
  month    =  {Mar},
  year     =  2018
}

@inproceedings{
schrittwieser2021reanalysis,
title={Online and Offline Reinforcement Learning by Planning with a Learned Model},
author={Julian Schrittwieser and Thomas K Hubert and Amol Mandhane and Mohammadamin Barekatain and Ioannis Antonoglou and David Silver},
booktitle={Advances in Neural Information Processing Systems},
year={2021},
url={https://openreview.net/forum?id=HKtsGW-lNbw}
}

@article{
wang2024goplan,
title={{GOP}lan: Goal-conditioned Offline Reinforcement Learning by Planning with Learned Models},
author={Mianchu Wang and Rui Yang and Xi Chen and Hao Sun and Meng Fang and Giovanni Montana},
journal={Transactions on Machine Learning Research},
issn={2835-8856},
year={2024},
url={https://openreview.net/forum?id=zOKAmm8R9B},
note={}
}

@InProceedings{oh2018selfimitation,
  title = 	 {Self-Imitation Learning},
  author =       {Oh, Junhyuk and Guo, Yijie and Singh, Satinder and Lee, Honglak},
  booktitle = 	 {Proceedings of the 35th International Conference on Machine Learning},
  pages = 	 {3878--3887},
  year = 	 {2018},
  volume = 	 {80},
  series = 	 {Proceedings of Machine Learning Research},
  month = 	 {Jul},
  publisher =    {PMLR},
  url = 	 {https://proceedings.mlr.press/v80/oh18b.html}
}

@ARTICLE{wang2023retroexplainer,
  title    = "Retrosynthesis prediction with an interpretable deep-learning
              framework based on molecular assembly tasks",
  author   = "Wang, Yu and Pang, Chao and Wang, Yuzhe and Jin, Junru and Zhang,
              Jingjie and Zeng, Xiangxiang and Su, Ran and Zou, Quan and Wei,
              Leyi",
  journal  = "Nature Communications",
  volume   =  14,
  number   =  1,
  pages    = "6155",
  month    =  {Oct},
  year     =  2023
}

@ARTICLE{hong2023egmcts,
  title    = "Retrosynthetic planning with experience-guided {Monte Carlo} tree
              search",
  author   = "Hong, Siqi and Zhuo, Hankz Hankui and Jin, Kebing and Shao, Guang
              and Zhou, Zhanwen",
  journal  = "Communications Chemistry",
  volume   =  6,
  number   =  1,
  pages    = "120",
  month    =  {Jun},
  year     =  2023
}

@ARTICLE{zhang2025dreamretroer,
  title    = "A data-driven group retrosynthesis planning model inspired by
              neurosymbolic programming",
  author   = "Zhang, Xuefeng and Lin, Haowei and Zhang, Muhan and Zhou, Yuan
              and Ma, Jianzhu",
  journal  = "Nature Communications",
  volume   =  16,
  number   =  1,
  pages    = "192",
  month    =  {Jan},
  year     =  2025
}

@article{maziarz2025syntheseus,
author ="Maziarz, Krzysztof and Tripp, Austin and Liu, Guoqing and Stanley, Megan and Xie, Shufang and Gaiński, Piotr and Seidl, Philipp and Segler, Marwin H. S.",
title  ="Re-evaluating retrosynthesis algorithms with {Syntheseus}",
journal  ="Faraday Discuss.",
year  ="2025",
volume  ="256",
issue  ="0",
pages  ="568-586",
publisher  ="The Royal Society of Chemistry",
doi  ="10.1039/D4FD00093E",
url  ="http://dx.doi.org/10.1039/D4FD00093E"}

@ARTICLE{sacha2021megan,
  title     = "Molecule Edit Graph Attention Network: Modeling Chemical
               Reactions as Sequences of Graph Edits",
  author    = "Sacha, Miko{\l}aj and B{\l}a{\.z}, Miko{\l}aj and Byrski, Piotr
               and D{\k a}browski-Tuma{\'n}ski, Pawe{\l} and Chromi{\'n}ski,
               Miko{\l}aj and Loska, Rafa{\l} and W{\l}odarczyk-Pruszy{\'n}ski,
               Pawe{\l} and Jastrz{\k e}bski, Stanis{\l}aw",
  journal   = "J. Chem. Inf. Model.",
  publisher = "American Chemical Society",
  volume    =  61,
  number    =  7,
  pages     = "3273--3284",
  month     =  {Jul},
  year      =  2021
}

@article{dong2021deep,
    author = {Dong, Jingxin and Zhao, Mingyi and Liu, Yuansheng and Su, Yansen and Zeng, Xiangxiang},
    title = {Deep learning in retrosynthesis planning: datasets, models and tools},
    journal = {Briefings in Bioinformatics},
    volume = {23},
    number = {1},
    pages = {bbab391},
    year = {2021},
    month = {Sep},
    issn = {1477-4054},
    doi = {10.1093/bib/bbab391},
    url = {https://doi.org/10.1093/bib/bbab391},
    eprint = {https://academic.oup.com/bib/article-pdf/23/1/bbab391/42230678/bbab391.pdf},
}

@ARTICLE{chen2021localretro,
  title     = "Deep Retrosynthetic Reaction Prediction using Local Reactivity
               and Global Attention",
  author    = "Chen, Shuan and Jung, Yousung",
  journal   = "JACS Au",
  publisher = "American Chemical Society",
  volume    =  1,
  number    =  10,
  pages     = "1612--1620",
  month     =  {Oct},
  year      =  2021
}

@ARTICLE{zhao2024meea,
  title    = "Efficient retrosynthetic planning with {MCTS} exploration
              enhanced {A}* search",
  author   = "Zhao, Dengwei and Tu, Shikui and Xu, Lei",
  journal  = "Communications Chemistry",
  volume   =  7,
  number   =  1,
  pages    = "52",
  month    =  {Mar},
  year     =  2024
}

@article{roucairol2024comparing,
author = {Roucairol, Milo and Cazenave, Tristan},
title = {Comparing search algorithms on the retrosynthesis problem},
journal = {Molecular Informatics},
volume = {43},
number = {7},
pages = {e202300259},
doi = {https://doi.org/10.1002/minf.202300259},
url = {https://onlinelibrary.wiley.com/doi/abs/10.1002/minf.202300259},
eprint = {https://onlinelibrary.wiley.com/doi/pdf/10.1002/minf.202300259},
year = {2024}
}

@inproceedings{yu2022grasp,
 author = {Yu, Yemin and Wei, Ying and Kuang, Kun and Huang, Zhengxing and Yao, Huaxiu and Wu, Fei},
 booktitle = {Advances in Neural Information Processing Systems},
 pages = {10257--10268},
 publisher = {Curran Associates, Inc.},
 title = {GRASP: Navigating Retrosynthetic Planning with Goal-driven Policy},
 url = {https://proceedings.neurips.cc/paper_files/paper/2022/file/42beaab8aa8da1c77581609a61eced93-Paper-Conference.pdf},
 volume = {35},
 year = {2022}
}

@ARTICLE{coley2017retrosim,
  title     = "Computer-assisted Retrosynthesis Based on Molecular Similarity",
  author    = "Coley, Connor W and Rogers, Luke and Green, William H and
               Jensen, Klavs F",
  journal   = "ACS Cent. Sci.",
  publisher = "American Chemical Society",
  volume    =  3,
  number    =  12,
  pages     = "1237--1245",
  month     =  {Dec},
  year      =  2017
}

@ARTICLE{tu2022graph2smiles,
  title     = "Permutation Invariant {Graph-to-Sequence} Model for
               Template-Free Retrosynthesis and Reaction Prediction",
  author    = "Tu, Zhengkai and Coley, Connor W",
  journal   = "J. Chem. Inf. Model.",
  publisher = "American Chemical Society",
  volume    =  62,
  number    =  15,
  pages     = "3503--3513",
  month     =  {Aug},
  year      =  2022
}

@misc{peng2019awr,
      title={Advantage-Weighted Regression: Simple and Scalable Off-Policy Reinforcement Learning}, 
      author={Xue Bin Peng and Aviral Kumar and Grace Zhang and Sergey Levine},
      year={2019},
      eprint={1910.00177},
      archivePrefix={arXiv},
      primaryClass={cs.LG},
      url={https://arxiv.org/abs/1910.00177}, 
}

@misc{nair2021awac,
      title={{AWAC}: Accelerating Online Reinforcement Learning with Offline Datasets}, 
      author={Ashvin Nair and Abhishek Gupta and Murtaza Dalal and Sergey Levine},
      year={2021},
      eprint={2006.09359},
      archivePrefix={arXiv},
      primaryClass={cs.LG},
      url={https://arxiv.org/abs/2006.09359}, 
}

@inproceedings{wang2018marwil,
 author = {Wang, Qing and Xiong, Jiechao and Han, Lei and sun, peng and Liu, Han and Zhang, Tong},
 booktitle = {Advances in Neural Information Processing Systems},
 pages = {},
 publisher = {Curran Associates, Inc.},
 title = {Exponentially Weighted Imitation Learning for Batched Historical Data},
 url = {https://proceedings.neurips.cc/paper_files/paper/2018/file/4aec1b3435c52abbdf8334ea0e7141e0-Paper.pdf},
 volume = {31},
 year = {2018}
}

@inproceedings{
wang2025lom,
title={Learning on One Mode: Addressing Multi-modality in Offline Reinforcement Learning},
author={Mianchu Wang and Yue Jin and Giovanni Montana},
booktitle={The Thirteenth International Conference on Learning Representations},
year={2025},
url={https://openreview.net/forum?id=upkxzurnLC}
}

@InProceedings{mao2023str,
  title = 	 {Supported Trust Region Optimization for Offline Reinforcement Learning},
  author =       {Mao, Yixiu and Zhang, Hongchang and Chen, Chen and Xu, Yi and Ji, Xiangyang},
  booktitle = 	 {Proceedings of the 40th International Conference on Machine Learning},
  pages = 	 {23829--23851},
  year = 	 {2023},
  volume = 	 {202},
  series = 	 {Proceedings of Machine Learning Research},
  month = 	 {Jul},
  publisher =    {PMLR},
  url = 	 {https://proceedings.mlr.press/v202/mao23c.html}
}

@inproceedings{
siegel2020abm,
title={Keep Doing What Worked: Behavior Modelling Priors for Offline Reinforcement Learning},
author={Noah Siegel and Jost Tobias Springenberg and Felix Berkenkamp and Abbas Abdolmaleki and Michael Neunert and Thomas Lampe and Roland Hafner and Nicolas Heess and Martin Riedmiller},
booktitle={International Conference on Learning Representations},
year={2020},
url={https://openreview.net/forum?id=rke7geHtwH}
}

@ARTICLE{yang2019dmpnn,
  title     = "Analyzing Learned Molecular Representations for Property
               Prediction",
  author    = "Yang, Kevin and Swanson, Kyle and Jin, Wengong and Coley, Connor
               and Eiden, Philipp and Gao, Hua and Guzman-Perez, Angel and
               Hopper, Timothy and Kelley, Brian and Mathea, Miriam and Palmer,
               Andrew and Settels, Volker and Jaakkola, Tommi and Jensen, Klavs
               and Barzilay, Regina",
  journal   = "J. Chem. Inf. Model.",
  publisher = "American Chemical Society",
  volume    =  59,
  number    =  8,
  pages     = "3370--3388",
  month     =  {Aug},
  year      =  2019
}

@ARTICLE{ruddigkeit2012gdb17,
  title     = "Enumeration of 166 Billion Organic Small Molecules in the
               Chemical Universe Database {GDB-17}",
  author    = "Ruddigkeit, Lars and van Deursen, Ruud and Blum, Lorenz C and
               Reymond, Jean-Louis",
  journal   = "J. Chem. Inf. Model.",
  publisher = "American Chemical Society",
  volume    =  52,
  number    =  11,
  pages     = "2864--2875",
  month     =  {Nov},
  year      =  2012
}

@article{zdrazil2023chembl,
    author = {Zdrazil, Barbara and Felix, Eloy and Hunter, Fiona and Manners, Emma J and Blackshaw, James and Corbett, Sybilla and de Veij, Marleen and Ioannidis, Harris and Lopez, David Mendez and Mosquera, Juan F and Magarinos, Maria Paula and Bosc, Nicolas and Arcila, Ricardo and Kizilören, Tevfik and Gaulton, Anna and Bento, A Patrícia and Adasme, Melissa F and Monecke, Peter and Landrum, Gregory A and Leach, Andrew R},
    title = {The ChEMBL Database in 2023: a drug discovery platform spanning multiple bioactivity data types and time periods},
    journal = {Nucleic Acids Research},
    volume = {52},
    number = {D1},
    pages = {D1180-D1192},
    year = {2023},
    month = {11},
    issn = {0305-1048},
    doi = {10.1093/nar/gkad1004},
    url = {https://doi.org/10.1093/nar/gkad1004},
    eprint = {https://academic.oup.com/nar/article-pdf/52/D1/D1180/55040046/gkad1004.pdf},
}

@article{liu2024retrocaptioner,
    author = {Liu, Xiaoyi and Ai, Chengwei and Yang, Hongpeng and Dong, Ruihan and Tang, Jijun and Zheng, Shuangjia and Guo, Fei},
    title = {RetroCaptioner: beyond attention in end-to-end retrosynthesis transformer via contrastively captioned learnable graph representation},
    journal = {Bioinformatics},
    volume = {40},
    number = {9},
    pages = {btae561},
    year = {2024},
    month = {Sep},
    issn = {1367-4811},
    doi = {10.1093/bioinformatics/btae561},
    url = {https://doi.org/10.1093/bioinformatics/btae561},
    eprint = {https://academic.oup.com/bioinformatics/article-pdf/40/9/btae561/60195134/btae561.pdf},
}

\clearpage

\appendix

\section{Case Studies} \label{sec:case_studies}
\begin{figure}[ht]
    \centering
    \includegraphics[width=0.95\linewidth]{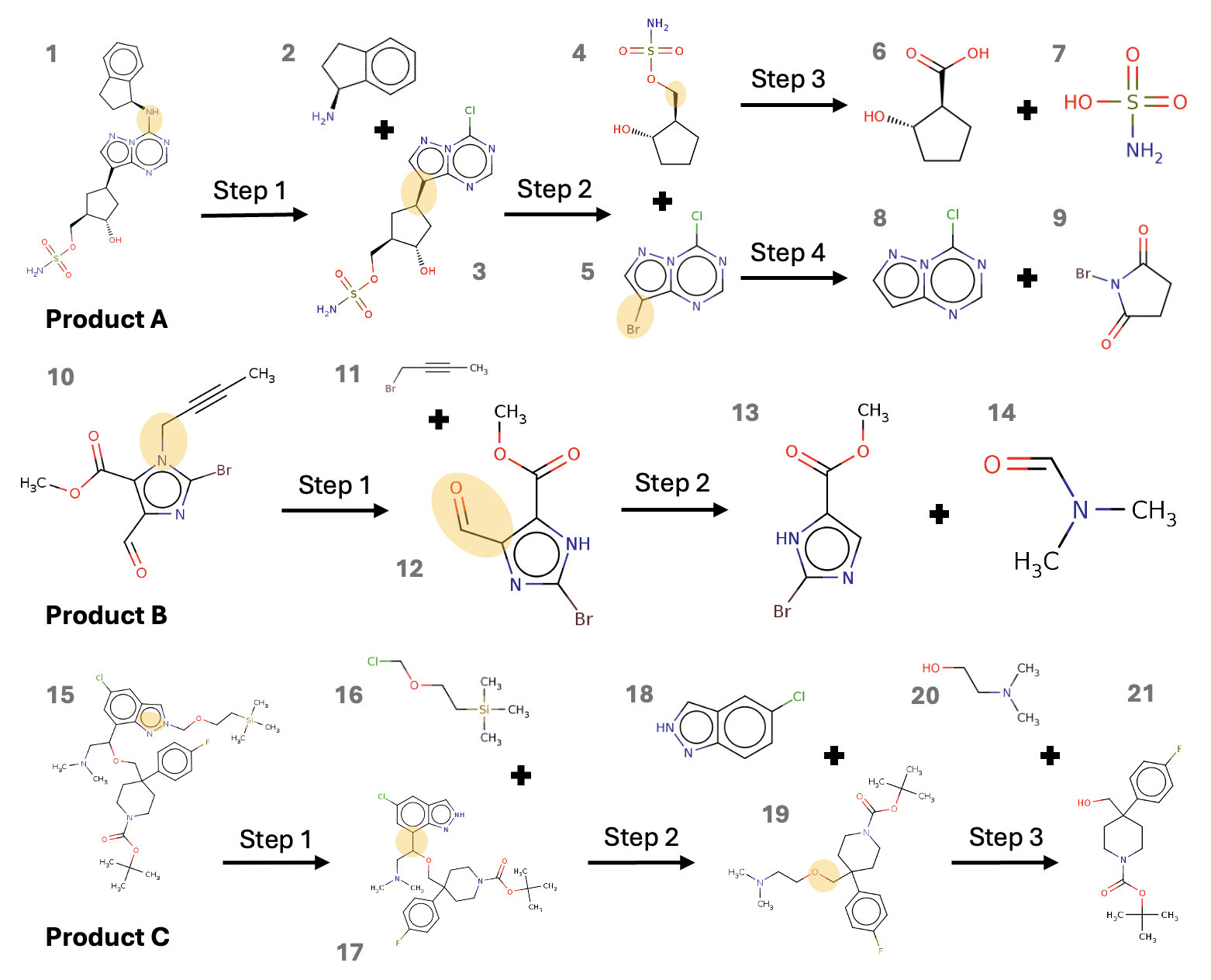}
    \caption{Predicted synthetic routes on three randomly selected molecules in Retro*-190. The yellow circles highlight the reaction centres. }
    \label{fig:case_studies}
\end{figure}

In Figure \ref{fig:case_studies}, we randomly select three targets from the Retro*‑190 benchmark and illustrate their complete synthetic trees as directly generated by InterRetro in a single forward pass.
(i) For the sulfonylated chiral cyclohexanol derivative, InterRetro proposes a four‑step route involving: a sulfonylation of a chiral alcohol to introduce the sulfonate ester, an electrophilic bromination of the diazine core, a C(sp$^3$)–C(sp$^2$) cross‑coupling to build the fused triazolopyrimidine system, and an N‑arylation to complete the final structure. 
(ii) For the alkyne‑substituted diazine, InterRetro constructs the target via two steps: first an acylation of the brominated azine to give the N‑formyl intermediate, followed by an SN2 N‑propargylation. 
(iii) For the complex triazolopyridine‑fused drug‑like scaffold, InterRetro suggests a three‑step route: a benzylic etherification that assembles the protected piperidine core, a C(sp³)–C(sp²) cross‑coupling to append the triazolopyridine fragment, and an N‑alkylation that installs the silyl‑protected side chain.
Across all three cases, InterRetro applies feasible chemical transitions and terminates in commercially available building blocks. A full list of the synthetic routes on Retro*‑190 is attached in the supplementary materials.

\section{Proofs and Further Theory}
\label{app:proofs}

\subsection{Proof of Proposition \ref{prop:q_and_v}}

\begin{proof} \label{proof:q_and_v}
    
Recall that 
\[
Q^\pi(s,a)
\;=\;
\mathbb{E}_{\tau \sim \pi}\!\Bigl[
\min_{p \in P(\tau)} \sum_{t=0}^T \gamma^t \, r\bigl(s_t\bigr)
\;\Big\vert\;
s_0 = s,\; a_0 = a
\Bigr].
\]
Since the transition function \(\mathcal{T}\) is deterministic, taking action \(a\) in \(s\) immediately leads to the set of next states \(\mathcal{T}(s,a)\). Observe that \(r(s)\) is \(1\) if and only if \(s\) is a building block (in which case no further transitions occur), and \(0\) otherwise. We prove the desired equality by a simple case distinction:

1. Case \(\boldsymbol{r(s)=1}\) (i.e., \(s\) is a building block).  
   In this case, once we reach \(s\), the path terminates and collects reward \(1\). Hence
   \[
   Q^\pi(s,a)
   \;=\;
   \underbrace{r(s)}_{=1}
   \;=\;
   r(s) + \gamma \,\bigl(1 - r(s)\bigr)
   \,\min_{s' \in \mathcal{T}(s,a)}\,V^\pi(s')
   \quad
   \text{(since \(1-r(s)=0\)).}
   \]

2. Case \(\boldsymbol{r(s)=0}\) (i.e., \(s\) is not a building block).  
   Any path extending from \((s,a)\) must proceed into one of the next states \(s' \in \mathcal{T}(s,a)\). Because we are taking a “worst-path” (minimum-return) perspective, the worst-case continuation value from \(s\) under action \(a\) is
   \[
   \min_{s' \in \mathcal{T}(s,a)} V^\pi(s'),
   \]
   and each step is discounted by \(\gamma\). Therefore,
   \[
   Q^\pi(s,a)
   \;=\;
   \underbrace{r(s)}_{=0}
   \;+\;
   \gamma 
   \,\min_{s' \in \mathcal{T}(s,a)}\,V^\pi(s')
   \;=\;
   r(s)
   \;+\;
   \gamma\,\bigl(1 - r(s)\bigr)
   \,\min_{s' \in \mathcal{T}(s,a)}\,V^\pi(s').
   \]

Combining both cases completes the proof:

\[
Q^\pi(s, a) 
\;=\;
r(s) 
\;+\; 
\gamma\,\bigl(1 - r(s)\bigr)\,\min_{s' \in \mathcal{T}(s,a)} V^\pi(s').
\]
\end{proof}

\subsection{Proof of Proposition \ref{prop:value_recursive}}
\begin{proof} \label{proof:value_recursive}
Recall that 
\[
V^\pi(s)
\;=\;
\mathbb{E}_{\tau \sim \pi}\!\Bigl[
\min_{p \in P(\tau)} \sum_{t=0}^T \gamma^t \,r\bigl(s_t\bigr)
\;\Big\vert\;
s_0 = s
\Bigr].
\]
When we are in state \(s\), the next action \(a\) is sampled from the policy \(\pi(\cdot\mid s)\). Because the environment is deterministic, taking \((s,a)\) leads to the set of next states \(\mathcal{T}(s,a)\). By the definition of the worst-path criterion (the inner minimum), we have
\[
Q^\pi(s,a)
\;=\;
r(s)
\;+\;
\gamma\,(1-r(s))\;\min_{s' \in \mathcal{T}(s,a)}\,V^\pi(s'),
\]
and
\[
V^\pi(s)
\;=\;
\sum_{a \in \mathcal{A}}\,\pi(a \mid s)\;Q^\pi(s,a).
\]
Substitute \(Q^\pi(s,a)\) into the sum:
\[
V^\pi(s)
\;=\;
\sum_{a \in \mathcal{A}} \pi(a \mid s)\Bigl[
  r(s) \;+\; 
  \gamma\,(1-r(s))\;\min_{s' \in \mathcal{T}(s,a)} V^\pi(s')
\Bigr].
\]
Since \(\sum_{a\in\mathcal{A}} \pi(a\mid s)=1\), we can factor out \(r(s)\) to obtain
\[
V^\pi(s)
\;=\;
r(s)
\;\underbrace{\sum_{a \in \mathcal{A}} \pi(a \mid s)}_{=1}
\;+\;
\gamma\,(1-r(s))\,
\sum_{a \in \mathcal{A}} \pi(a \mid s)\,\min_{\,s' \in \mathcal{T}(s,a)} V^\pi(s').
\]
Hence,
\[
V^\pi(s)
\;=\;
r(s)
\;+\;
\gamma\,\bigl(1-r(s)\bigr)
\sum_{a \in \mathcal{A}} \pi(a \mid s)\,\min_{\,s' \in \mathcal{T}(s,a)} V^\pi(s'),
\]
completing the proof.
\end{proof}

\subsection{Proof of Proposition \ref{prop:bellman_optimality}}
\label{appendix:optimality_and_contraction}

We consider the space $\mathcal{V}$ of all bounded real-valued functions $V: \mathcal{S} \rightarrow \mathbb{R}$ defined over the state space $\mathcal{S}$. For any function $V \in \mathcal{V}$, its $L_\infty$-norm (or max-norm) is given by $\|V\|_\infty = \sup_{s \in \mathcal{S}} |V(s)|$. The space $(\mathcal{V}, \|\cdot\|_\infty)$ is a complete metric space (a Banach space).

The Bellman optimality operator $B^*: \mathcal{V} \rightarrow \mathcal{V}$ for the worst-path objective, with discount factor $\gamma \in (0,1)$, is defined for any value function $V \in \mathcal{V}$ and any state $s \in \mathcal{S}$ as:
\begin{equation} \label{eq:appendix_bellman_optimality_operator}
(B^*V)(s) = r(s) + \gamma (1 - r(s)) \max_{a \in \mathcal{A}} \left[ \min_{s' \in \mathcal{T}(s, a)} V(s') \right].
\end{equation}
Note that if $s \in \mathcal{S}_{bb}$, then $r(s)=1$, so $1-r(s)=0$, and $(B^*V)(s) = r(s)$. If $s \notin \mathcal{S}_{bb}$, then $r(s)=0$, so $1-r(s)=1$, and $(B^*V)(s) = \gamma \max_{a \in \mathcal{A}} \left[ \min_{s' \in \mathcal{T}(s, a)} V(s') \right]$. The optimal value function $V^*$ is the unique fixed point of this operator, i.e., $V^* = B^*V^*$.

\begin{proposition} \label{prop:contraction_mapping_appendix}
The Bellman optimality operator $B^*$ defined in Eq.~\eqref{eq:appendix_bellman_optimality_operator} is a $\gamma$-contraction mapping with respect to the $L_\infty$-norm. That is, for any $V_1, V_2 \in \mathcal{V}$:
\[ \|B^*V_1 - B^*V_2\|_\infty \leq \gamma \|V_1 - V_2\|_\infty. \]
\end{proposition}

\begin{proof} \label{proof:contraction_mapping_appendix}
Let $V_1, V_2 \in \mathcal{V}$ be two arbitrary bounded value functions. We consider any state $s \in \mathcal{S}$.

Case 1: $s \in \mathcal{S}_{bb}$.
In this scenario, $(B^*V_1)(s) = r(s)$ and $(B^*V_2)(s) = r(s)$, as the second term in Eq.~\eqref{eq:appendix_bellman_optimality_operator} (involving the maximisation) vanishes because $1-r(s)=0$. Thus, $|(B^*V_1)(s) - (B^*V_2)(s)| = 0$. The contraction inequality $0 \leq \gamma \|V_1 - V_2\|_\infty$ therefore holds trivially.

Case 2: $s \notin \mathcal{S}_{bb}$.
In this case, $r(s)=0$, so $1-r(s)=1$.
Then,
\[ (B^*V_1)(s) = \gamma \max_{a \in \mathcal{A}} \left[ \min_{s' \in \mathcal{T}(s, a)} V_1(s') \right], \]
\[ (B^*V_2)(s) = \gamma \max_{a \in \mathcal{A}} \left[ \min_{s' \in \mathcal{T}(s, a)} V_2(s') \right]. \]
Therefore,
\[ |(B^*V_1)(s) - (B^*V_2)(s)| = \gamma \left| \max_{a \in \mathcal{A}} \left[ \min_{s' \in \mathcal{T}(s, a)} V_1(s') \right] - \max_{a \in \mathcal{A}} \left[ \min_{s' \in \mathcal{T}(s, a)} V_2(s') \right] \right|. \]
Let $f_V(a) = \min_{s' \in \mathcal{T}(s,a)} V(s')$. The expression is $\gamma |\max_a f_{V_1}(a) - \max_a f_{V_2}(a)|$.
Using the property that for any functions $g_1, g_2$, $|\max_x g_1(x) - \max_x g_2(x)| \leq \sup_x |g_1(x) - g_2(x)|$, we have:
\[ \left| \max_{a \in \mathcal{A}} f_{V_1}(a) - \max_{a \in \mathcal{A}} f_{V_2}(a) \right| \leq \sup_{a \in \mathcal{A}} |f_{V_1}(a) - f_{V_2}(a)| \]
\[ = \sup_{a \in \mathcal{A}} \left| \min_{s' \in \mathcal{T}(s,a)} V_1(s') - \min_{s' \in \mathcal{T}(s,a)} V_2(s') \right|. \]
Using the property that for any functions $h_1, h_2$, $|\min_y h_1(y) - \min_y h_2(y)| \leq \sup_y |h_1(y) - h_2(y)|$:
\[ \left| \min_{s' \in \mathcal{T}(s,a)} V_1(s') - \min_{s' \in \mathcal{T}(s,a)} V_2(s') \right| \leq \sup_{s' \in \mathcal{T}(s,a)} |V_1(s') - V_2(s')|. \]
Combining these,
\begin{align*}
|(B^*V_1)(s) - (B^*V_2)(s)| &\leq \gamma \sup_{a \in \mathcal{A}} \left[ \sup_{s' \in \mathcal{T}(s,a)} |V_1(s') - V_2(s')| \right] \\
&\leq \gamma \sup_{s'' \in \mathcal{S}} |V_1(s'') - V_2(s'')| \quad (\text{as } \mathcal{T}(s,a) \subseteq \mathcal{S} \text{ and we take sup over } a) \\
&= \gamma \|V_1 - V_2\|_\infty.
\end{align*}
This inequality holds for any $s \notin \mathcal{S}_{bb}$.

Combining Case 1 and Case 2, for all $s \in \mathcal{S}$:
\[ |(B^*V_1)(s) - (B^*V_2)(s)| \leq \gamma \|V_1 - V_2\|_\infty. \]
Taking the supremum over all $s \in \mathcal{S}$ on the left side:
\[ \|B^*V_1 - B^*V_2\|_\infty = \sup_{s \in \mathcal{S}} |(B^*V_1)(s) - (B^*V_2)(s)| \leq \gamma \|V_1 - V_2\|_\infty. \]
Since $0 < \gamma < 1$, $B^*$ is a $\gamma$-contraction mapping in $(\mathcal{V}, \|\cdot\|_\infty)$. The space $\mathcal{V}$ of bounded real-valued functions on $\mathcal{S}$, equipped with the $L_\infty$-norm, is a complete metric space (a Banach space). Therefore, by the Banach fixed-point theorem, $B^*$ has a unique fixed point $V^* \in \mathcal{V}$ such that $V^* = B^*V^*$.
Furthermore, for any initial bounded value function $V_0 \in \mathcal{V}$, the sequence $V_{k+1} = B^*V_k$ (i.e., value iteration) converges to $V^*$. The existence of a unique $V^*$ implies the existence of at least one stationary deterministic optimal policy $\pi^*$ such that for $s \notin \mathcal{S}_{bb}$:
\[ \pi^*(s) \in \arg\max_{a \in \mathcal{A}} \left[ \min_{s' \in \mathcal{T}(s, a)} V^*(s') \right]. \]
\end{proof}

\subsection{Proof of Proposition \ref{prop:policy_improvement}}
 \label{proof:policy_improvement}

\begin{proof}
We begin by recalling a standard identity from policy improvement theory, which relates the value difference between two policies $\pi^{i+1}$ and $\pi^i$ to the expected advantage under $\pi^{i+1}$:
\[
V^{\pi^{i+1}}(s) - V^{\pi^i}(s) = \mathbb{E}_{a \sim \pi^{i+1}(\cdot \mid s)} \left[ A^{\pi^i}(s, a) \right],
\]
where $A^{\pi^i}(s, a) = Q^{\pi^i}(s, a) - V^{\pi^i}(s)$ is the advantage function under the behaviour policy $\pi^i$.

The update rule defined by the weighted imitation objective yields a new policy $\pi^{i+1}$ that is proportional to the exponentiated advantage:
\[
\pi^{i+1}(a \mid s) = \frac{\pi^i(a \mid s) \cdot \exp(\beta A^{\pi^i}(s, a))}{Z(s)},
\]
where $Z(s) = \sum_{a'} \pi^i(a' \mid s) \cdot \exp(\beta A^{\pi^i}(s, a'))$ is a normalising constant. This update ensures that $\pi^{i+1}$ remains in the same support set as $\pi^i$.

Substituting into the value difference identity, we get:
\[
V^{\pi^{i+1}}(s) - V^{\pi^i}(s) = \sum_a \pi^{i+1}(a \mid s) \cdot A^{\pi^i}(s, a) = \frac{\sum_a \pi^i(a \mid s) \cdot \exp(\beta A^{\pi^i}(s, a)) \cdot A^{\pi^i}(s, a)}{\sum_a \pi^i(a \mid s) \cdot \exp(\beta A^{\pi^i}(s, a))}.
\]

This expression is a weighted average of the advantages $A^{\pi^i}(s,a)$, where the weights are strictly positive and increasing in the advantage. Therefore, unless all advantages are exactly zero, the expectation is strictly non-negative:
\[
V^{\pi^{i+1}}(s) \geq V^{\pi^i}(s), \quad \forall s \in \mathcal{S}.
\]
\end{proof}

\section{Graph2Edits as Single-step Model} \label{app:graph2edits}

Graph2Edits \cite{zhong2023graph2edits} considers one‑step retrosynthesis as a graph‑editing game.  Starting from the product molecule, the model autoregressively predicts a short sequence of primitive edits — Delete Bond, Change Bond, Change Atom, Attach Leaving Group, and finally a Terminate token.  At each step the current intermediate graph is embedded by a directed message‑passing neural network (D‑MPNN), whose atom‑ and bond‑level embeddings are fed to three linear heads that score every possible bond edit, atom edit and the termination symbol.  The highest‑scoring edit is applied to yield the next intermediate, and the process repeats until Terminate is chosen, at which point the intermediates have been fully converted to the predicted reactants.  This edit vocabulary, mined from training data, covers $99.9\%$ of USPTO‑50k reactions and encodes stereochemistry as well as functional leaving groups. 

Trained with teacher forcing, the system already achieves a $55\%$ top‑1 exact‑match accuracy on USPTO‑50k and maintains high validity and diversity even for long or stereochemically rich reactions; these strengths make it an ideal single‑step model inside our multi‑step InterRetro framework. 

\end{document}